\def\year{2022}\relax
\documentclass[letterpaper]{article} % DO NOT CHANGE THIS
\usepackage{aaai22}  % DO NOT CHANGE THIS
\usepackage{times}  % DO NOT CHANGE THIS
\usepackage{helvet}  % DO NOT CHANGE THIS
\usepackage{courier}  % DO NOT CHANGE THIS
\usepackage[hyphens]{url}  % DO NOT CHANGE THIS
\usepackage{graphicx} % DO NOT CHANGE THIS
\usepackage{CJKutf8}
\usepackage{hyperref}
\usepackage{natbib}  % DO NOT CHANGE THIS AND DO NOT ADD ANY OPTIONS TO IT
\usepackage{caption} % DO NOT CHANGE THIS AND DO NOT ADD ANY OPTIONS TO IT
\DeclareCaptionStyle{ruled}{labelfont=normalfont,labelsep=colon,strut=off} % DO NOT CHANGE THIS
\usepackage{algorithm}
\usepackage{algorithmic}
\usepackage{latexsym}
\usepackage{amsmath}
\usepackage{booktabs}
\usepackage{url}
\usepackage{arydshln}
\usepackage{comment}
\usepackage{multirow}
\usepackage{color}
\usepackage{subfigure}
\usepackage{newfloat}
\usepackage{listings}
\floatstyle{ruled}
\newfloat{listing}{tb}{lst}{}
\floatname{listing}{Listing}
\title{Automatic Product Copywriting for E-Commerce}
\author {
    Xueying Zhang\textsuperscript{\rm 1}\equalcontrib,
    Yanyan Zou\textsuperscript{\rm 2}\equalcontrib, 
    Hainan Zhang\textsuperscript{\rm 2}, 
    Jing Zhou\textsuperscript{\rm 2},
    Shiliang Diao\textsuperscript{\rm 2}, 
    Jiajia Chen\textsuperscript{\rm 2}, 
    Zhuoye Ding\textsuperscript{\rm 2}, 
    Zhen He\textsuperscript{\rm 2}, 
    Xueqi He\textsuperscript{\rm 2},
    Yun Xiao\textsuperscript{\rm 1}, 
    Bo Long\textsuperscript{\rm 2}, 
    Han Yu\textsuperscript{\rm 3$\dagger$}, 
    Lingfei Wu\textsuperscript{\rm 1$\dagger$}
}
\affiliations {
    \textsuperscript{\rm 1} JD.COM Silicon Valley Research Center\\
    \textsuperscript{\rm 2} JD.COM\\
    \textsuperscript{\rm 3} Nanyang Technological University\\
    \textsuperscript{$\dagger$}Corresponding authors: han.yu@ntu.edu.sg, lingfei.wu@jd.com
}
\usepackage{bibentry}
\begin{document}

\maketitle

\begin{abstract}
Product copywriting is a critical component of e-commerce recommendation platforms. It aims to attract users' interest and improve user experience by highlighting product characteristics with textual descriptions.
In this paper, we report our experience deploying the proposed Automatic Product Copywriting Generation (APCG) system into the JD.com e-commerce product recommendation platform. It consists of two main components: 1)
natural language generation, which is built from a transformer-pointer network and a pre-trained sequence-to-sequence model based on millions of training data from our in-house platform; and 2)
copywriting quality control, which is based on both automatic evaluation and human screening. 
For selected domains, the models are trained and updated daily with the updated training data.
In addition, the model is also used as a real-time writing assistant tool on our live broadcast platform.
The APCG system has been deployed in JD.com since Feb 2021. By Sep 2021, it has generated 2.53 million product descriptions, and improved the overall averaged click-through rate (CTR) and the Conversion Rate (CVR) by 4.22\% and 3.61\%, compared to baselines, respectively on a year-on-year basis.
The accumulated Gross Merchandise Volume (GMV) made by our system is improved by 213.42\%, compared to the number in Feb 2021.
\end{abstract}

\section{Introduction} 

\begin{figure*}[t!]
\centering
\includegraphics[width=1\linewidth]{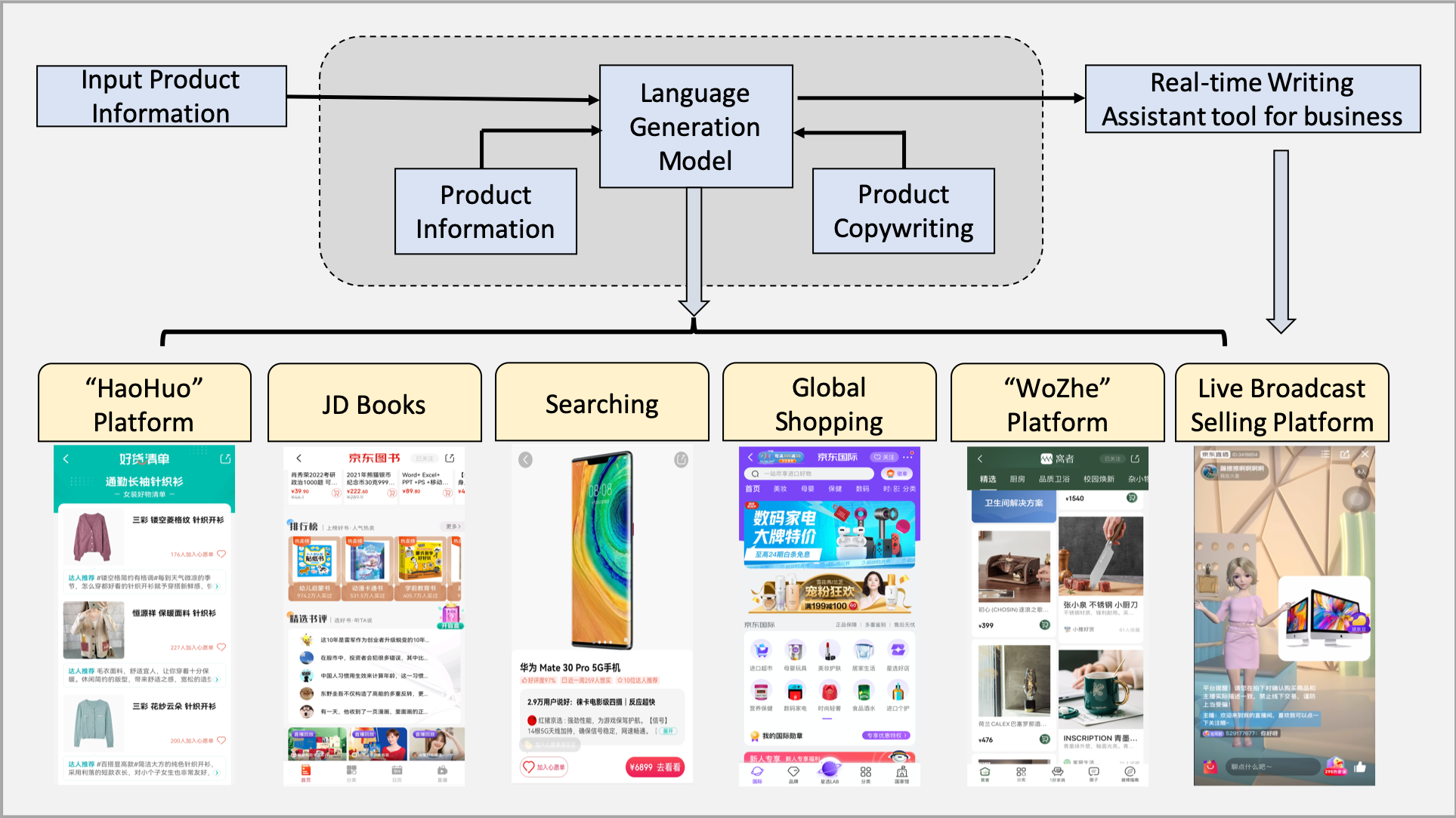}
\caption{Overview of the proposed Automatic Product Copywriting Generation (APCG) system.}
\label{fig1}
\end{figure*}

With the rapid development of the Internet, online shopping has become indispensable in people's daily life. 
E-commerce is now a mainstay way of shopping for many people all over the world.
Product advertising copywriting plays an important role in e-commerce platforms, where well-written advertisements can help attract customers and, in turn, increase sales.
Traditionally, product copywriting is performed by human copywriters. 
This approach has important limitations. Firstly, the efficiency of human copywriters cannot match the growth rate of new products. Secondly, 
manpower cost increases as the number of products in the system increases. Last but not least, specialist training and tutoring is required for copywriters for different products and scenarios. 
To overcome these issues, automatic product copywriting generation has become an important line of research in e-commerce.

Natural language generation (i.e., NLG) research focuses on the construction of computer systems that can produce understandable texts in human languages from some underlying non-linguistic or textual representation of information \cite{reiter_dale_2000}. 
NLG systems combine knowledge from both the language and the application domain to automatically generate desired texts.
In recent years, NLG techniques have been developing rapidly and applied to a wide range of applications. 
For instance, abstract text summarization \cite{rush2015neural,michele2000headline,kevin2000statistics} generates a summarized version of input text to highlighted certain information; 
dialogue generation \cite{ijcai2018-614,zhang2020dialogpt} generates responses according to the previous chats; 
machine translation \cite{bahdanau2016neural} translates the input texts into a different language; and
storytelling or poem generation \cite{Yao_Peng_Weischedel_Knight_Zhao_Yan_2019,li-etal-2018-generating-classical} generates creative texts.

In the early stage of text generation research, due to limitation of data and computational resources, models relied heavily on feature engineering. 
In this case, features are extracted from raw data based on human knowledge and preference, and fed into the models to learn from the limited data \cite{laffertyCrf,och-etal-2004-smorgasbord}. 
With the development of deep learning, features are learned automatically and defined implicitly during model training. 
The natural language generation problem has been framed as a sequence-to-sequence task and various studies have been proposed to enhance the model architecture,  
such as convolutional neural networks (CNN) \cite{gehring2017convolutional}, recurrent neural networks (RNN) \cite{sutskever2014sequence},
graph neural networks (GNN) \cite{wu2020comprehensive}, and attention-based architecture \cite{bahdanau2016neural,vaswani2017attention} (especially the transformer network~\cite{vaswani2017attention}). 

Recently, the pre-training plus fine-tuning paradigm has gained traction and been widely applied in real-world applications. 
Under this paradigm, models are first pre-trained on large-scale corpus \cite{devlin-etal-2019-bert,brown2020language}, or on domain-specific corpus \cite{zhang2021dsgpt} with well-defined pre-training tasks. 
The resulting pre-trained models are then fine-tuned to adapt to different downstream tasks. 
Research in this paradigm mainly focuses on defining training objectives in the pre-training and fine-tuning stages. 
With increasing computational resources and the emerging transformer architecture~\cite{vaswani2017attention}, recent studies (e.g., GPT2 \cite{radford2019language}, CTRL \cite{keskar2019ctrl}, UNiLM \cite{dong2019unified}) utilizing pre-training and fine-tuning techniques have achieved outstanding performance in language generation tasks. 

In this paper, we report on our experience deploying the proposed Automatic Product Copywriting Generation (APCG) system into the JD.com\footnote{\url{https://www.jd.com/}} e-commerce product recommendation platform. It is made up of two main components: 1) a
natural language generation (NLG) module, which is built from a transformer-pointer network and a pre-trained sequence-to-sequence model; and 2)
a copywriting quality control module, which involves both automatic evaluation and human screening to ensure the quality of the generated descriptions. 
For selected domains, the models are trained and updated daily with the updated training data.
In addition, the model is also used as a real-time writing assistant tool on our live broadcast platform.

Since its deployment in JD.com in Feb 2021, APCG has generated 2.53 million product descriptions by Sep 2021. As a result, the overall click-through rate (CTR) has been improved by 4.22\%, the conversion rate (CVR) has been improved by 3.61\%, and the accumulated gross merchandise volume (GMV) has more than doubled on a year-on-year basis.

\section{Application Description}

An online e-commerce website typically describes a product $p$ from several aspects, including the product title $t$, a set of the product attribute $A$ and the corresponding attribute values $V$ for each attribute $a\in A$, as well as the short advertisement slogan $s$ written by advertising experts.
The product title $t$ usually describes the product with a short text, expressing the key information, such as the brand name and product type.
The product attribute set $A$ captures the properties of a product from different perspectives. For example, the product functions, intended users, styles, colors and materials. 
The corresponding values of the attributes from $A$ form the product attribute value set $V$.
For example, the product function feature can be \emph{breathable, resistant}, and the intended user feature can be \emph{ladies}.
Given the title $t$, attribute set $A$, the corresponding attribute value set $V$ and the advertisement slogan $s$, the task of product copywriting is to generate a well-written description $D$ that is able to present product characteristics for the e-commerce system to attract user interest and keep them informed of the key features of the product quickly.  
An example of product copywriting is demonstrated in Figure \ref{fig4}.

%There has been some related studies that also work on product copywriting generation.
Early work on product copywriting applied template-based generation methods with statistical knowledge extracted from the training data~ \cite{wang2017statistical}, while the template coverage and diversity are limited by the training corpus.
\citeauthor{chen2019towards}~\citeyear{chen2019towards} designed a transformer-based generation model to incorporate the external knowledge (i.e., Wikipedia knowledge base).
Besides, the textual information (i.e., product title and attributes) and the product image features are also incorporated. 
The performance of such a model heavily relies on the availability of related knowledge. 
Similarly, \citeauthor{li2020aspect}~\citeyear{li2020aspect} constructed a set of aspects (i.e., salient attributes) and the corresponding keywords for each aspect, which, however, requires expert annotations.
An adaptive posterior Transformer-based network \cite{zhan2021probing} has been designed to utilize relevant information from customer reviews extracted by an adaptive posterior distillation module, where the product title and attribute information is considered as the main model input.

However, neither the external knowledge base nor the customer reviews are always accessible. 
In practice, for the e-commerce platform, newly released products are emerging every day, where the product information is quite limited.
In addition, such kind of external features may contain noise, especially for newly released products.
This reduces the quality of the generated product descriptions.
On the other hand, each product is accompanied with several short advertising phrases written by experts. However, existing studies did not leverage this information when generating product descriptions.
In this work, we jointly consider the product textual features (i.e., title, attribute and advertising slogan) as product input information, and develop a transformer-pointer model and a pre-trained sequence-to-sequence model for automatic product copywriting in e-commerce settings. Both of such two models constitute our Automatic Product Copywriting Generation (APCG) system.

\begin{figure}[b!]
\centering
\includegraphics[width=1\columnwidth]{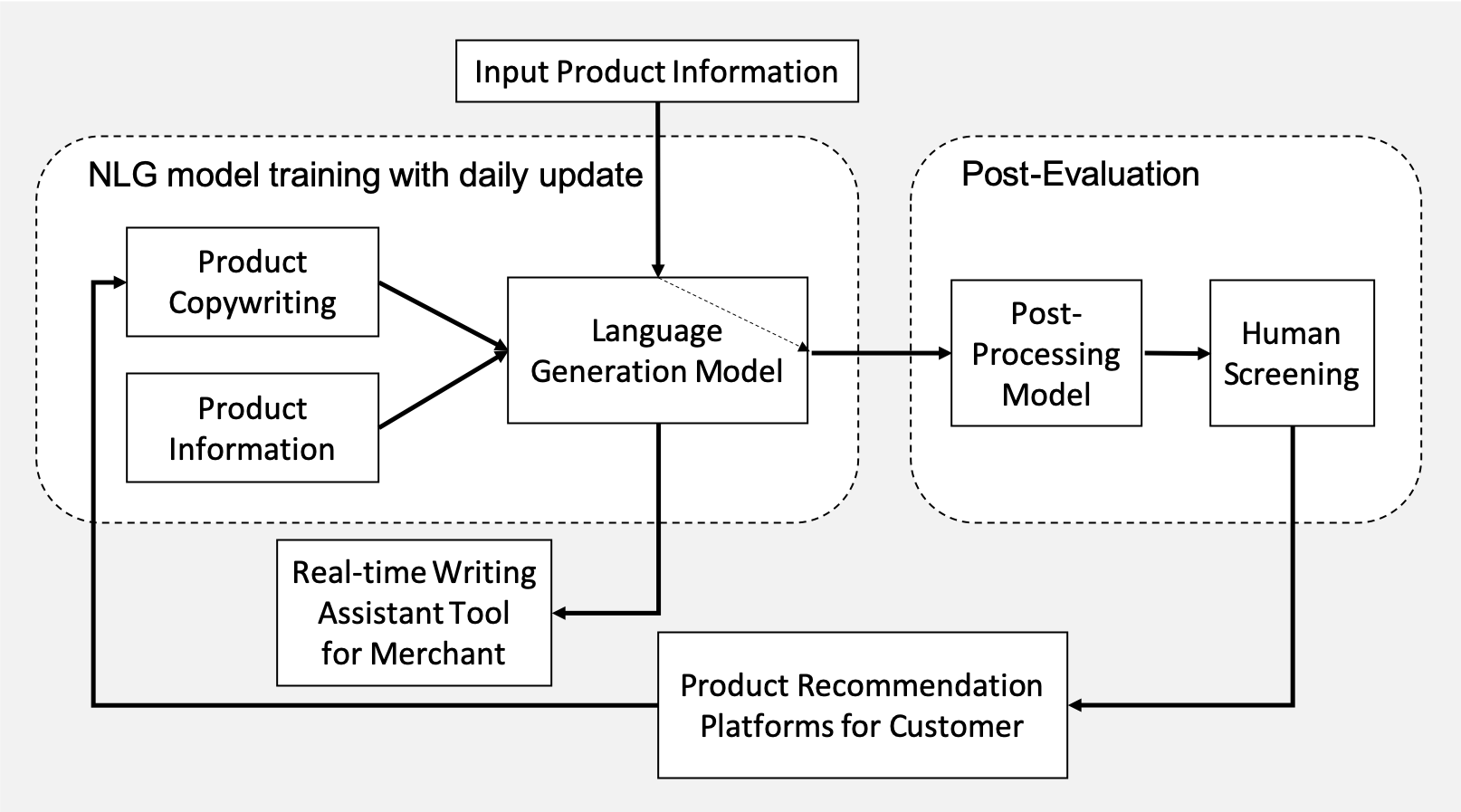} % Reduce the figure size so that it is slightly narrower than the column. Don't use precise values for figure width.This setup will avoid overfull boxes.
\caption{Workflow of the APCG system.}
\label{fig2}
\end{figure}

As illustrated in Figure \ref{fig1}, our proposed APCG system is designed to serve a variety of applications in the JD.com e-commerce platform, including the ``HaoHuo'' (referring to ``Discovery Goods'') product recommendation platform, JD books, searching, global shopping, the ``WoZhe'' platform (mainly for home appliance) and the live broadcast selling platform, etc.
As shown in Figure \ref{fig2}, the overall system consists of the following steps:
\begin{itemize}
    \item [1)] NLG model training: To collect data, including both product information and descriptions, for training the NLG model. Different models are selected upon different scenarios. Both the transformer-pointer network and the pre-trained language models are applied in this stage.
    \item [2)] Post-evaluation for direct customer use: Post-processing models are utilized here to filter out contents which do not meet the requirements. Human screening is performed after model-based filtering. The contents produced here are directly used by the product recommendation platform for JD customers.
    \item [3)] Real-time writing assistant tool for sellers: Besides generating and filtering contents based on proposed product skus, a real-time writing assistant tool based on APCG has been developed for JD sellers. With this tool, JD sellers can input their desired products, and it will automatically generate product descriptions. Sellers can then edit the generated contents, and display them to customers.
    \item [4)] Automatic procedure with daily update: APCG is trained daily to meet the latest writing preference based newly updated data.
\end{itemize}

\section{Use of AI Technology}
In this section, we describe the key techniques adopted by the APCG AI Engine.

\subsection{Data Collection and Cleaning} 

\subsubsection{Data Collection}

We built a large-scale real-world dataset from our in-house platform which includes two parts: 1) product description data, and 2) product information data. 
The product descriptions are written by professional copywriters with good knowledge of marketing. For each product, the description includes both advertising contents and the title.
For the product information data, we collected multiple types of data from our in-house database, including product titles, attributes, product detail images and user reviews.

\subsubsection{Data Cleaning}
After obtaining the large-scale datasets, we perform filtering and cleaning on the raw data to construct our training dataset.
The product descriptions are filtered by human evaluation and rule-based methods, and used as the outputs of our model. The inputs include information from product titles, attributes, product detail images and user reviews. For each category, we select different groups of attributes which have high relevance with the target products. For product detail images, we leverage the optical character recognition (OCR) and classification techniques to extract key information about the product. Firstly, texts are extracted from the product detail images. Then, the extracted texts are ranked in descending order of their importance and relevance using the classification model. Finally, highly ranked texts are selected and merged as the final OCR input. We use the ranking model to select useful reviews, and a summarization model to obtain highlights of selected reviews, which are then used as additional inputs.  

\subsection{The Transformer-Pointer Network} 

Based on this large-scale proprietary dataset, the automatic product copywriting problem can be regarded as an abstractive summarization problem.
We designed a pointer-generator network, which allows the transformer network to generate new words and copy words form the input text at the same time. Through this transformer-pointer network, issues of inaccurate descriptions and unknown words can be handled since both extractive and abstractive approaches are executed simultaneously. 

\begin{figure}[t!]
\centering
\includegraphics[width=1.0\columnwidth]{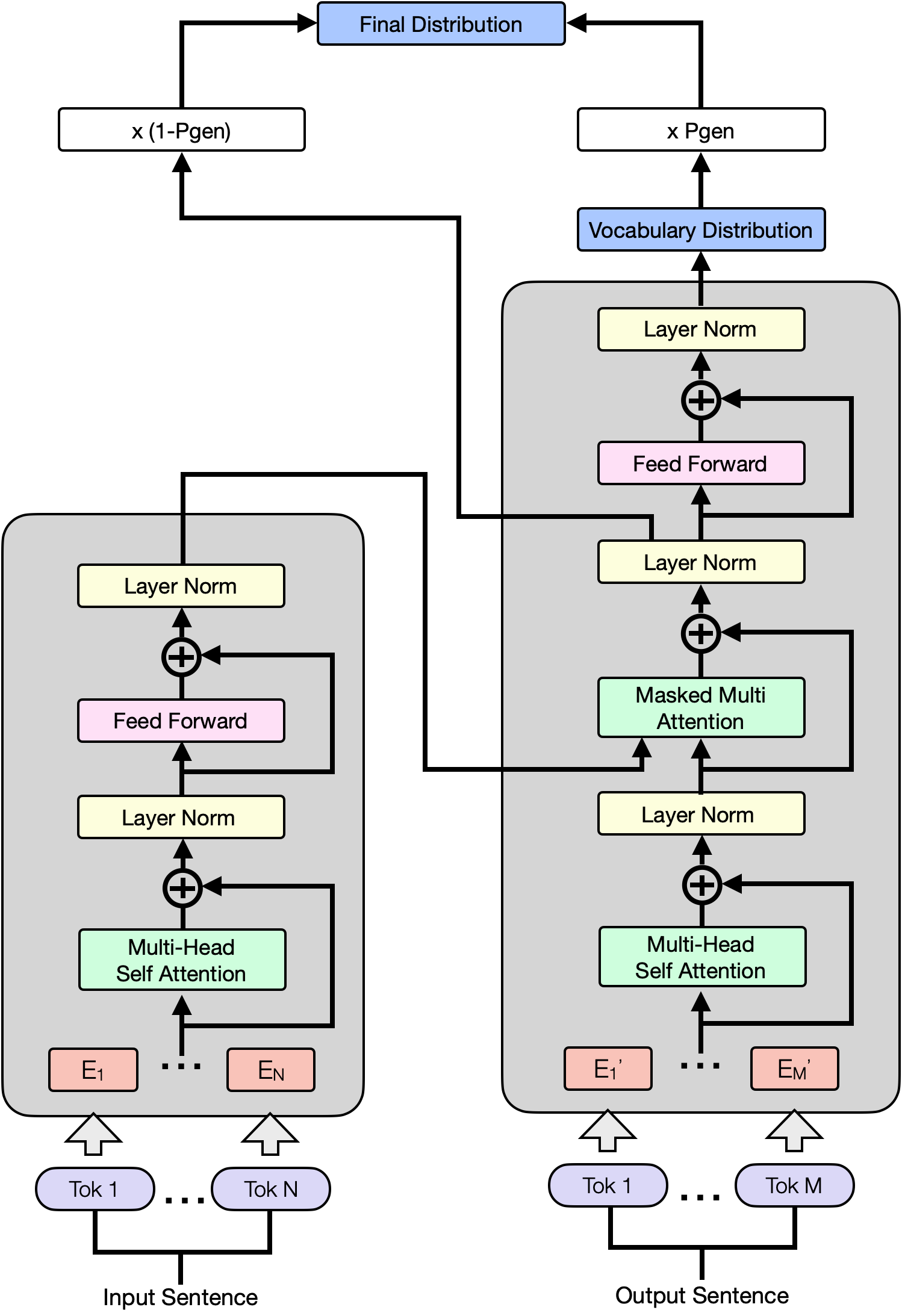} % Reduce the figure size so that it is slightly narrower than the column. Don't use precise values for figure width.This setup will avoid overfull boxes.
\caption{The transformer-pointer network structure.}
\label{fig3}
\end{figure}

The proposed transformer-pointer network is a combination of a pointer-generator network and a transformer structure. 
As shown in Figure \ref{fig3}, by introducing the trainable probability $P_{gen}$, the model learns to choose whether to use words from the vocabulary source or from the input text. 
APCG will select a new word from the vocabulary distributions with a probability of $P_{gen}$, and copy words directly from the input text through the selective pointer with a probability of $(1-P_{gen})$. 
We take the encoder-decoder attentions in the last decoder layer as the copy distribution. Note that for the multi-head attention, we obtain the copy distributions by summing across multiple heads.

\subsection{The Pre-trained Language Model}
Training a neural network from scratch requires large amounts of data with high-quality labels, especially true for the generation task. Unlabeled open-domain data (i.e., without gold standard annotations) are readily available from online sources such as Wikipedia.
Recently, pre-trained models, such as OpenAI GPT \cite{radfordimproving,radford:2019:arxiv}, BERT \cite{devlin2019bert}, RoBERTa \cite{liu2019roberta} and BART \cite{lewis2019bart}, have become popular for NLP tasks.
They have significantly improved the state-of-the-art performance on various natural language understanding (NLU) tasks (e.g., SQuAD question answering \cite{rajpurkar2016squad}, natural language inference \cite{bowman2015large},
text classification \cite{zhang2015character}) and natural language generation tasks (e.g., machine translation \cite{song2019mass} and text summarization \cite{zou2020pre}). This demonstrates the effectiveness and transferability of models pre-trained on large scale unlabeled data.
Recent research progresses in \cite{brown2020language} have demonstrated that pre-trained models can play as an important role in few-shot learning with limited training instances.

In the product copywriting scenario, there are different categories of products (e.g., clothing, mobile phones, sportswear).
The product characteristics and data properties vary significantly across different categories.
One straightforward solution is to train the transformer-pointer model on the training data for each category individually. However, this will incur high computation and maintenance costs.
In addition, as the products in the e-commerce are updated frequently, there is often limited or even no training data for newly released products in practice.
Therefore, we adopt the pre-trained model for product copywriting.

\begin{comment}
 \cite{dong2019unified} (UniLM) proposed a unified language model that can be used for both natural language understanding and generation tasks, which is pre-trained using masked, unidirectional and sequence-to-sequence language modeling objectives.
 \cite{song2019mass} (MASS) proposed a method to pre-train a sequence-to-sequence Transformer by masking a span of text and then predicting the masked tokens.
Their pre-training task is similar to our MDG task, but we apply a different masking strategy and predict the original text.
%	Besides, we propose another two tasks for {\sc seq2seq} model pre-training. 
 \citet{song2019mass} tested their model on sentence-level tasks (e.g., machine translation and sentence compression), while we aim to solve document-level tasks (e.g., abstractive document summarization). 
 \citet{lewis2019bart} (BART) adopted the combination of text infilling and sentence permutation as a single objective for sequence-to-sequence Transformer pre-training.
Differently, we propose three objectives and use them individually. Specifically, MDG replaced each selected token with a masked token in the input sequence.
 \citet{raffel2019exploring} (T5) studies different pre-training objectives, model architectures, and unlabeled datasets. 
ProphetNet \cite{yan2020prophetnet} predicts the next $n$ tokens simultaneously.
 \citet{zhang2019pegasus} (PEGASUS) proposed to remove/mask sentences from an input document and learn to generate such removed/masked sentences for pre-training.
\end{comment}

We adopt the sequence-to-sequence transformer network \cite{vaswani2017attention}, which comprises of an encoder and a decoder, as the backbone architecture.
Given an input token sequence $X= (x_1, x_2, \dots, x_{|X|})$ paired with its corresponding output sequence $Y=(y_1, y_2, \dots, y_{|Y|})$, we aim to learn the model parameters $\theta$ and estimate the conditional probability:
\begin{equation}
\vspace{-0mm}
P(Y|X;\theta) = \prod_{t=1}^{|Y|} p(y_t|y_{<t};X;\theta)
\end{equation}
where $y_{<t}$ stands for all tokens before position $t$ (i.e., $y_{<t}=(y_1, y_2, \dots, y_{t-1})$).
Given the whole training set $(\mathcal{X}, \mathcal{Y})$, this model can be trained by maximizing the log-likelihood of the training input-output pairs:
\begin{equation}
\vspace{-0mm}
\mathcal{L}(\theta; \mathcal{X}, \mathcal{Y}) = \sum_{(X, Y)\in(\mathcal{X}, \mathcal{Y})} \log P(Y|X;\theta).
\end{equation}
We pre-train the model using sequence-to-sequence objectives. 

Recall that the attributes of a product are represented by the set $A$ without ordering information. 
We observe that the transformer model generates different outputs when receiving attributes in different orders as inputs. This can sometimes produce poor quality outputs.
On the other hand, given the product title, attributes and advertising slogans, the task of product copywriting is to generate the description, where the generated text contains tokens copied from input text and novel words and phrases that are not contained in the input. 
Based on these observations, we design two sequence-to-sequence objectives for pre-training.

\paragraph{Sentence Re-ordering}
Inspired by \cite{zou2020pre}, we incorporate the sentence re-ordering objective (SR) in APCG.
We first divide an unlabeled document into multiple sentences based on full stop signs.
%	Let us change the notation of a document slightly in this paragraph. Let $X=(S_1, S_2, \dots, S_m)$ denote a document, where $S_i = (w^i_1, w^i_2, \dots, w^i_{|S_i|})$ is a sentence, $w^i_j$ is $j^{\text{th}}$ word in $S_i$ and $m$ is the number of sentences.
For better demonstration, we change the notation of a document slightly in this paragraph. Let $X=(S_1 || S_2 || \dots || S_m)$ denote a document, consisting of several sentences, where $S_i$ is a sentence, $m$ is the number of sentences, and $||$ refers to sentence concatenation.
%	 $X$ is still a sequence of tokens (by concatenating tokens in all sentences). 
The sentence index order in $X$ can be represented as $\mathcal{O}=(1, 2, \dots, m)$. 
We then shuffle the sentences in the document.
In other words, the items in the order $\mathcal{O}$ are re-arranged and we obtain a shuffled order $\mathcal{O}_{S}=(a_1,a_2,\dots, a_m)$, where $a_i$ refers to sentence index, $1 \leq a_i \leq m$,  $1 \leq a_j \leq m$, and $a_i \neq a_j$ for any $i, j \in [1, m]$ and $i\neq j$.
% 	A permutation function $A=\text{\tt permutation}(m)$ is defined to rearrange the sentence order and generate a new order $(a_1,a_2,\dots, a_m)$.
Concatenating sentences following $\mathcal{O}_{S}$, we obtain a \emph{shuffled} document $\hat{X}_S=(S_{a_1} || S_{a_2} || \dots || S_{a_m})$.
%	Let $A=\text{\tt permutation}(m)=(a_1,a_2,\dots, a_m)$ to denote a permuted range of $(1, 2, \dots, m)$ and therefore $\hat{X}_S=(S_{a_1}, S_{a_2}, \dots, S_{a_m})$ is the \emph{shuffled} document.
%	 Note that $\hat{X}_S$ is a sequence of tokens by concatenating all \emph{shuffled} sentences. 
The sequence-to-sequence model takes the shuffled document $ \hat{X}_S$ as input and is pre-trained to re-instate the original one $X$.
%	can be trained on $\langle \hat{X}_S, X \rangle$ pairs constructed from unlabeled documents, as demonstrated in Figure \ref{fig:architecture}.
The training objective is expressed as:
\begin{equation*}
\vspace{-0mm}
\mathcal{L}(\theta; \mathcal{X}) = \sum_{X\in\mathcal{X}} \log P(X| \hat{X}_S;\theta).
\end{equation*}

\paragraph{Pseudo Summary Generation}
This objective is adopted from \cite{zhuiyit5pegasus}.
Suppose a document is denoted as $X=(S_1 || S_2 || \dots || S_m)$.
We select around $m/4$ sentences from the document $X$, and concatenate into $\hat{X}_{M}$.
The remaining $3m/4$ sentences are concatenated into $X_{R}$.
The sentences in $\hat{X}_{M}$ are carefully selected so that $\hat{X}_{M}$ and $X_R$ contain the longest subsequences among all alternatives. 
The longer text $X_R$ is considered as the model input, while the shorter text $\hat{X}_{M}$ is regarded as the model output.
Thus, the training objective is expressed as:
\begin{equation*}
\vspace{-0mm}
\mathcal{L}(\theta; \mathcal{X}) = \sum_{X\in\mathcal{X}} \log P(X_R| \hat{X}_{M};\theta).
\end{equation*}
After a sequence-to-sequence model is pre-trained, we fine tune the model on the parallel copywriting training data, consisting of instances from all categories.

% Based on above observations and inspired by \cite{devlin2019bert}, we design a masking policy with three strategies.
% Within the selected subsequence $S$, each token is processed as follows: 1) with probability of 80\%, the token is replaced with the symbol \texttt{[MASK]}; 2) with 10\% of the time, the token is replaced with a random token sampled from the source dictionary; 3) for the remaining 10\% cases, the token remains unchanged.
\begin{comment}
During pre-training, we consider two settings. \textbf{Setting one:} pre-training a model with one single objective, i.e., SR, NSG or MDG, resulting in three different pre-trained models.
\textbf{Setting two}: employing all three objectives.
For each training batch, we randomly choose one objective and each objective is used for $1/3$ of the training time, obtaining one model (i.e., ALL, see Section \ref{sec:exp}).

For better reference, we name our model as STEP (i.e., \textbf{s}equence-\textbf{t}o-s\textbf{e}quence \textbf{p}re-training) that can be used to denote a sequence-to-sequence model pre-trained using our proposed objective(s).
\end{comment}

\begin{CJK*}{UTF8}{gbsn}
\subsection{Post-Processing}

Although the proposed language generation models can generate texts with good quality most of the time, there is still possibility that the generated descriptions might not be accurate. Thus, we incorporate post-processing models into APCG to further enhance copywriting quality to meet the high expectation in practical e-commerce applications. 

In some cases, the descriptions generated by the model are not consistent with product information. For example, for iPhone 11, the model generates the product description of ``iPhone 11采用升降式摄像头'' (iPhone 11 with a foldable webcam), which is not an actual feature of iPhone 11. Although cases like this are rare, they can negatively affect user experience. To overcome this issue, we designed a rule-based product-word and number checking model. We collect words and numbers related for each product item, and construct our specific e-commerce domain specific word dictionary. We develop rule-based models for each product category based on these dictionaries to filter out generated descriptions with inaccurate information. 
The rules are based on combinations of terms, where single words within them are not affected. 

Although NLG models are trained to achieve good fluency and grammar correctness, they are not fool proof, especially when the input information is too new for the model. To address this limitation, we train grammar checking models and leverage the Adaboost algorithm to deploy these models to filter out generated descriptions with poor grammar quality.
These post-processing approaches help APCG filter out poorly generated description and enhance the copywriting quality.

\end{CJK*}

\section{Application Development and Deployment}

\subsection{Offline Evaluation}
We compare our transformer-pointer model (TP) and pre-training based copywriting generation model (PCG) with two baselines before making deployment decisions:
\begin{itemize}
\item The template-based generation model (i.e., Template) \cite{wang2017statistical}; and
\item The attribute mining based generation model (i.e., A-Mining) \cite{li2020aspect}.
\end{itemize}

For fair comparison, we randomly selected 500 products from different categories, and generated descriptions using the four models. 
We used SacreBLEU \cite{post2018call}, ROUGE \cite{lin2004rouge}, BLEU \cite{papineni2002bleu}, and Meteor \cite{lavie2004significance} to measure the quality of different generation model outputs.
We report the SacreBLUE, ROUGE-1, ROUGE-2, ROUGE-L, BLEU-1, BLEU-2, BLUE-3, BLEU-4 and Meteor scores in Table \ref{tab:eval_auto}. The larger the values of these metrics, the better the performance of a model. It can be observed that PCG consistently achieves the best performance under all evaluation metrics.
%	The ROUGE scores are computed using the \texttt{ROUGE-1.5.5.pl} script\footnote{https://github.com/bheinzerling/pyrouge.git}.

Since descriptions generated by models may produce incoherent or grammatically incorrect outputs, we also evaluated these four models by eliciting human judgements. 
%We compared our two models with human references (denoted as Human), as well as two baselines.
Ten human participants are presented with a product and a list of descriptions generated by the four different models in random orders.
Then, they are asked to rank these descriptions according to fluency (\emph{is the description grammatically sound?}), relevance (\emph{is the description relevant to the product?}), and diversity (\emph{does the text describe the product from different perspectives?})
The results are listed in Table \ref{tab:eval_human}. It can be observed that PCG achieves the best performance under all human evaluation metrics. These results helped us make the decision to incorporate TP and PCG into the APCG system.

\begin{table*}[t]
\begin{center}
\caption{Offline evaluation results for the four models.}
\label{tab:eval_auto}
\resizebox*{1\textwidth}{!}{
\begin{tabular}{lccccccccc}
			\toprule
Model                    & SacreBLUE & ROUGE-1 & ROUGE-2 & ROUGE-L & BLEU-1 & BLEU-2 & BLEU-3 & BLEU-4 & Meteor \\
\midrule
Template                     & 4.84      & 14.20    & 2.65     & 11.01    & 27.95   & 15.94   & 10.09   & 6.80    & 12.72  \\
A-Mining                    & 5.35      & 16.33    & 3.43     & 13.04    & 30.24   & 17.31   & 10.83   & 7.27    & 13.57  \\
\hdashline
TP & 5.81      & 17.37    & 4.14     & 14.06    & 29.80   & 17.90   & 11.70   & \textbf{8.17}    & 13.77  \\ 
PCG          & \textbf{6.19}      & \textbf{18.98}    & \textbf{4.37}     & \textbf{15.66}    & \textbf{30.29}   & \textbf{18.34}   & \textbf{11.97}   & \textbf{8.17}    & \textbf{14.04} \\ \bottomrule
\end{tabular}
}
\end{center}
\end{table*}

\begin{table}[t]
\begin{center}
\caption{Human evaluation results for the four models.}
\label{tab:eval_human}
\begin{tabular}{lccc}
\toprule
Model                    & Fluency & Relevance & Diversity \\\midrule
% Human & 1.78      & 1.89     & 1.79    \\
Template                        & 1.78      & 1.89     & 1.79         \\
A-Mining                        & 1.80      & 1.90     & 1.77     \\
\hdashline
TP 
& 1.80      & 1.84     & 1.81     \\
PCG         & \textbf{1.96}      & \textbf{1.91}     & \textbf{1.87}    \\ \bottomrule
\end{tabular}
\end{center}
\end{table}

\subsection{Deployment Usage}

The procedure of automatic product copywriting generation includes the following steps: 1) obtaining the required product skus from the database; 2) generating product descriptions and performing post-processing; 3) submitting the generated contents for human screening; 4) storing approved contents into the database for customer service provision. 
Daily uploads to the descriptions database can reach up to 28,570 generated descriptions, with an average daily acceptance rate of 80\%. 
The usage of the JD product recommendation platform generates new data daily, our NLG models are trained and updated daily with these new data.

\begin{figure}[t!]
\centering
\includegraphics[width=1.0\columnwidth]{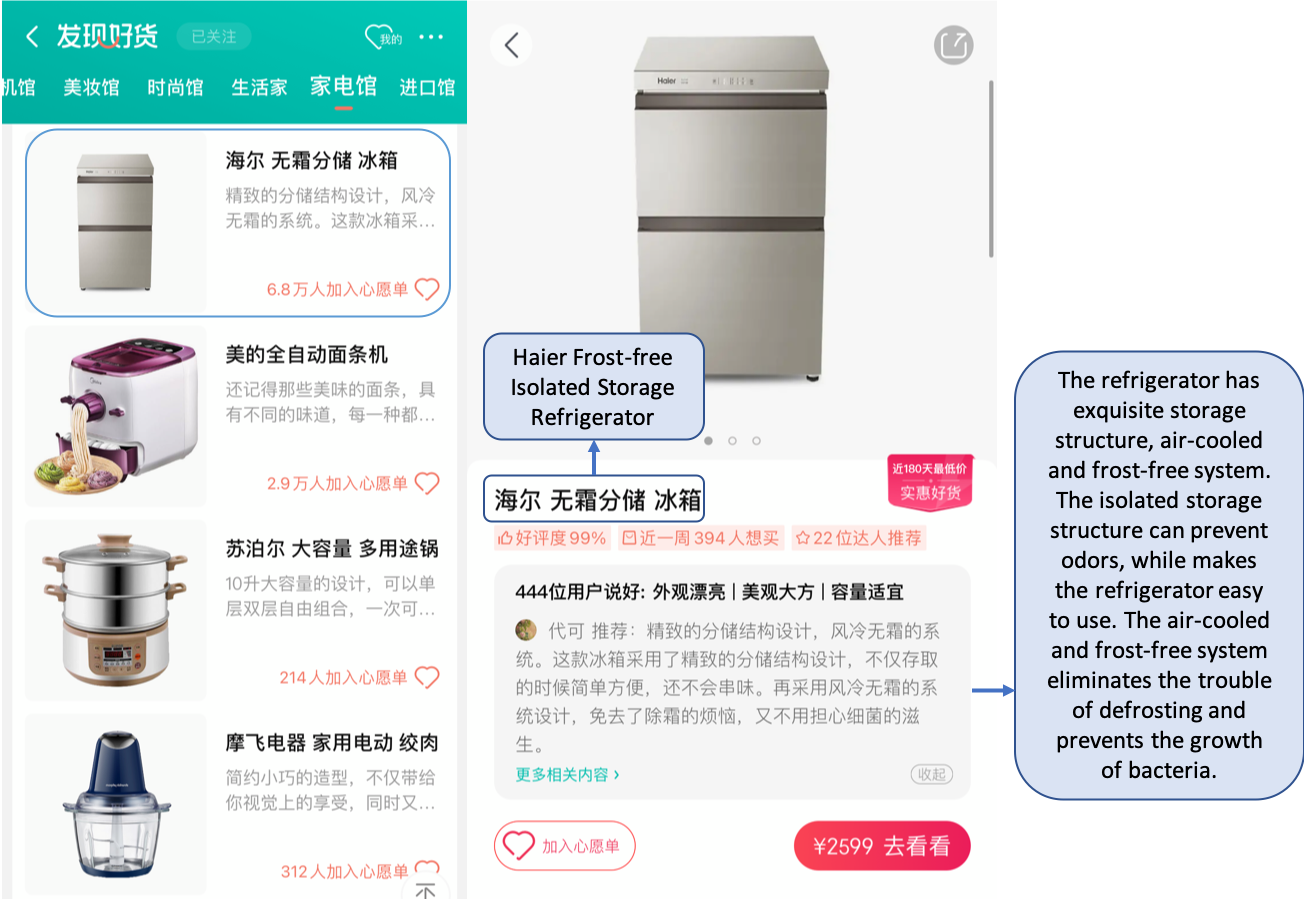} 
\caption{Examples of APCG generated descriptions on the JD e-commerce product recommendation platform.}
\label{fig4}
\end{figure}

\begin{figure}[t!]
\centering
\includegraphics[width=0.85\columnwidth]{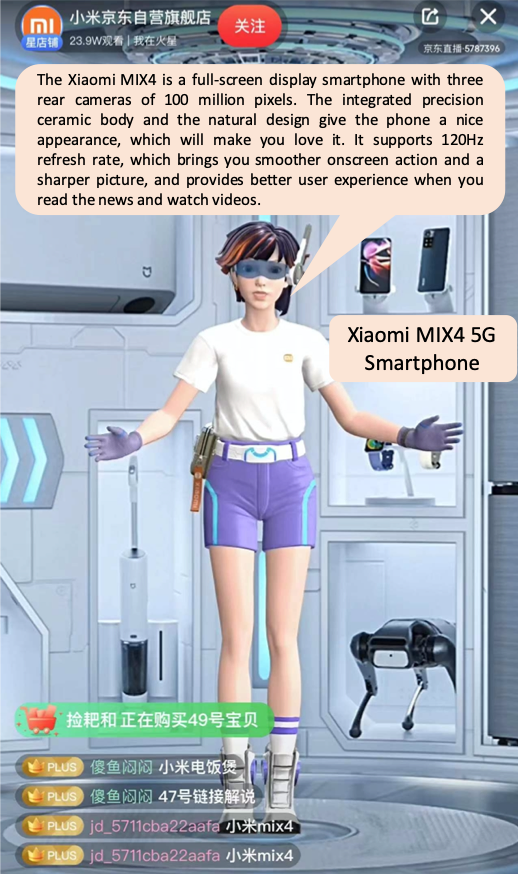} 
\caption{Examples of APCG generated descriptions on the JD live broadcast selling platform. The virtual salesperson reads the product descriptions in the virtual studio.}
\label{fig5}
\end{figure}

\begin{CJK*}{UTF8}{gbsn}
As illustrated in Figure \ref{fig2}, the deployed APCG system serves two main usage scenarios: 1) content generation for customers, and 2) real-time writing assistant for sellers. 
In the first scenario, the contents generated by APCG are stored in our database and displayed directly on the JD product recommendation platform.
Figure \ref{fig4} shows an example of the product recommendation platform for customers, ``HaoHuo''. The English translations of the generated texts are shown in the pop up bubbles in the figure.
The generated product descriptions, paired with a short title, is pushed to the Discovery Goods Channel in the JD e-commerce website. The short title consists of three phrases: 1) the brand name, 2) one of salient product characteristics, and 3) the product name, which make up a product headline.
To further ensure the user experience, we also display three images for a product.
Overall, the short title, description and three images are combined together to depict a product in the Discovery Goods Channel.
APCG only focuses on the description generation task.

In the second scenario, a real-time writing assistant tool is designed for sellers, especially for sellers promoting products on the live broadcast platform. 
The sellers can input the product ID they would like to advertise, and the APCG writing assistant tool will generate product descriptions fitting for the current context in real time. 
Figure \ref{fig5} shows the use case in the live broadcast selling platform.
By Sep, 2021, the APCG writing assistant tool has been adopted by more than 2,000 merchants with daily live streams up to 150 sessions.

\end{CJK*}

As real-time performance is important for the second use case, we optimize the design of APCG for this purpose. We split the encoder and decoder of our transformer model, and design interfaces to realize the beam search algorithm. The two main interfaces include the encoder predictor and decoder predictor. The encoder predictor receives the input tokens, and returns the token IDs and the encoded embedding. For the decoder predictor, the inputs include the token IDs, the encoded embedding and the current prefix. It outputs the top-$k$ candidates of the predicted indices, tokens and probabilities. Due to language generation problems, contextual words need to be generated one-by-one. Thus, when generating descriptions for one item, the encoder interface is called once and the decoder interface is called multiple times depending on the length of the context and the beam size. For beam search, extra storage is required for holding the candidates within the beam. 

We conduct latency test for our real-time writing assistant tool. Table \ref{table3} shows that the GPU predictor can handle high query request traffic of up to 100 queries per second (QPS), which is four times the throughput of the CPU predictor. The average latency and TP-99 (i.e., the minimum time under which 99\% of requests have been served) achieved by using GPU are about 1/5 of using CPU.

To further enhance the efficiency, before calling our language generation engine to generate contents in the real time, we first check whether the proposed input product is in our copywriting description database. If so, we will directly provide the previously generated contents stored in the database to save time.

\begin{table}[t]
\renewcommand\arraystretch{1.25}
\centering
\caption{Latency test comparison of CPU and GPU services.}
\label{table3}
\begin{tabular}{cccc}
\toprule
Service & QPS & avg latency (ms) & TP-99 (ms) \\ \midrule
CPU predictor & 23.0 & 80.0 & 106.0 \\
GPU predictor & 100.0 &	15.6 &	19.2 \\
\bottomrule
\end{tabular}
\end{table}

\section{Maintenance} 
The key performance indicator used to monitor the performance of APCG is the daily acceptance rate. If this rate becomes low, we will perform manual checks on the rejected descriptions and fix the issues in the model accordingly.
Besides this, no other major maintenance task has been required since its deployment in Feb 2021.

\section{Application Use and Payoff} 
%(Y)
The APCG system has been deployed in JD.com since Feb 2021. To demonstrate the payoff generated by APCG, we compare key performance indicators before and after deploying APCG on the JD Discover Goods Channel and its mobile app.
A standard A/B testing is used on such online platforms to evaluate the benefit of the APCG-generated product descriptions for the business on the platform. The impact on the business of the platform is measure by the Click-Through Rate (CTR) and Conversion Rate (CVR).
CTR is defined as:
\begin{equation}
    \text{CTR} = \frac{\text{\#clicked\_product}}{\text{\#pv\_product}}
\end{equation}
where click\_product is the number of clicks on the products, and pv\_product is the number of page views for the product shown to the users. 
CVR is defined as:
\begin{equation}
    \text{CVR} = \frac{\text{\#converted\_product}}{\text{\#clicked\_product}}
\end{equation}
where the converted\_product refers to the number of times a product has been purchased by users.

For online deployment, we randomly selected around 500,000 products covering 20 categories, and generated corresponding descriptions using both the Transformer-Pointer and Pre-trained models.
The baseline method is the A-Mining model which was used by the platform prior to the deployment of APCG.
We first conduct A/B testing for the baseline model and Transformer-Pointer model. 
During the online A/B testing period, over five million people visited the Discovery Good Channel, providing us with adequate opportunities for testing. 
The A/B testing system divided these visiting users into two equal-sized groups, and directed them into two individual buckets (i.e., buckets A and B).
For users in bucket A (i.e., the baseline group), the system shows them the product descriptions generated by the A-Mining model. 
For users in bucket B (i.e., the experiment group), the system shows them the product descriptions generated by the Transformer-Pointer model in APCG. 
%As show in Figure \ref{???}, the product copywrting is displayed on the front page within the Discovery Goods Channel, which is thus a crucial factor to determine whether a user would like to click the product and then make a purchase. 

The CTR and CVR for all products during the A/B testing period are calculated.
We found that the our Transformer-Pointer model significantly outperforms the previous method by a large margin.
Specifically, the TP model improved the CTR by 3.44\% and CVR by 3.40\% over the A-Mining model.
The increase of the CTR indicates that users prefer to click the products with descriptions generated by our system.
The improvement of CVR demonstrates that the descriptions generated by our model is successful in convincing users to make purchases.
%Therefore, we adopt the Transformer-Pointer model to generate copywriting for products and regularly update the descriptions as well as extend the coverage of the copywriting for the products.

Another A/B test has been conduced with our pre-trained model (PCG). This time, we use our Transformer-Pointer model as the baseline. All other A/B test settings remain the same.
Compared to the baseline, the CTR and CVR are improved by a relatively small margin of 1.3\% and 0.8\%, respectively, which is in line with expectation. 
Recall that, the Transformer-Pointer model is deployed individually for each category, while the PCG model is able to provide description generation services for all categories.
However, the pre-trained model incurs high computation costs. 
Thus, during APCG deployment, PCG mainly focuses on categories with limited training data or newly released products.

Between Feb 2021 and Sep 2021, APCG has generated 2.53 million product descriptions for the JD.com e-commerce platform. The overall averaged CTR achieved by APCG is 4.22\% higher than that achieved by the previous system year-on-year (yoy). The improvement in terms of CTR is 3.61\% yoy.
The accumulated Gross Merchandise Volume (GMV) made by our system is improved by 213.42\%, compared to the start number in Feb 2021.
By introducing automated copywriting, our daily uploads to the descriptions database can reach up to 28,570 generated descriptions, equivalent to the productivity of thousands of human experts, which significantly saves human labor costs.
Although the product copywriting task in the JD Discover Goods Channel is currently jointly performed by expert writers and the APCG systems,
we observed that expert writers often prefer to create descriptions for popular products. Differently, our system (especially the pre-trained model) is able to cover both popular and long-tail products, which helps build a healthy ecosystem. 
Thus, it is expected that the AI system will take over this task completely in the near future.

\section{Lessons Learned During Deployment}
At the time of submission of this paper, APCG has been successfully deployed for around six months and has led to significant positive impact on JD.com's business.
Several lessons we have learned during model deployment could be beneficial for other like-minded researchers and practitioners who wish to deploy cutting-edge AI technologies into real-world applications.

Firstly, besides the model capacity, the quality of training data is of paramount importance.
The cleaning procedures of raw data (e.g., removing poor samples from training set and specifying group of important attributes) play an important role in model development.
Today, data pre-processing is mainly based on human experiences and rules, which is expensive and might not work well in new domains. 
A possible alternative approach one can take is to design a classification model that can identify low quality data from the training set (e.g., through influence function-based approaches \cite{Chen-et-al:2021HyDRA}).

Secondly, the AI generated descriptions are not fool proof. Thus, in order to ensure the users can have a reasonably good experience, post-processing of AI generated descriptions in the production platform is necessary to filter out any inconsistent or low quality contents. Currently, this step is often based on rules and involves humans in the loop. Nevertheless, it is a necessary consideration during deployment to uphold user experience.
%Currently, we design several rules to filter out such bad outputs, rather that a control module planted within the generation model. However, how to make generation results more controllable is an remaining open questions in the community. 

%Secondly, notice that we have two lines of generation model, i.e., transformer-pointer models for each product category and pre-trained T5-pegasus model for all categories. The transformer-pointer model is light, which can be used to well serve for specific categories, while the pre-trained T5-pegasus model is able to transfer to new domains, especially there is limited data for training.  Combining both resulting a good deployment that is able to support different needs.

\section{Conclusions and Future Work} 
In this paper, we report our work on developing and deploying an automatic product copywriting generation system (APCG) for the JD e-commerce product recommendation platform. 
The proposed system consists of a transformer-pointer network and a pre-trained sequence-to-sequence model to perform natural language generation tasks in order to generate descriptions for products automatically. The system incorporates automatic evaluation and human screening modules to guard against the occasional low quality generated textual contents and ensure a high level of user experience. 
By adopting the APCG system, a high volume of 2.53 million product descriptions have been generated over a seven month period. The overall average CTR achieved by the JD Discover Goods Channel product recommendation platform is improved by 4.22\%, while its CVR is improved by 3.61\% on an year-on-year basis. 
The accumulated GMV made by our system is improved by 213.42\%, compared to the start number in Feb 2021.

In future work, we will focus on enhancing the model performance and expanding the application scenarios:
\begin{itemize}
\item Domain-specific pre-trained models: our current experience shows that public pre-trained models can help enhance the quality of natural language generation. We will develop pre-trained models specific for the e-commerce domain to further improve model performance. 
\item Expending application scenarios: besides product description generation, automatic natural language generation can benefit other application scenarios as well. We will explore applying APCG to other application scenarios such as personalized product description generation and multi-product advertisement post generation.
\end{itemize}

\section{Acknowledgments}
Han Yu is supported by the National Research Foundation, Singapore under its AI Singapore Programme (AISG Award No: AISG2-RP-2020-019); the Nanyang Assistant Professorship (NAP); and the RIE 2020 Advanced Manufacturing and Engineering (AME) Programmatic Fund (No. A20G8b0102), Singapore. Any opinions, findings and conclusions or recommendations expressed in this material are those of the author(s) and do not reflect the views of National Research Foundation, Singapore.
% Use \bibliography{yourbibfile} instead or the References section will not appear in your paper
% \nobibliography{aaai22}
\bibliography{IAAI_JDCopywriting}

\begin{thebibliography}{41}
\providecommand{\natexlab}[1]{#1}

\bibitem[{Bahdanau, Cho, and Bengio(2016)}]{bahdanau2016neural}
Bahdanau, D.; Cho, K.; and Bengio, Y. 2016.
\newblock Neural Machine Translation by Jointly Learning to Align and
  Translate.
\newblock arXiv:1409.0473.

\bibitem[{Banko, Mittal, and Witbrock(2000)}]{michele2000headline}
Banko, M.; Mittal, V.~O.; and Witbrock, M.~J. 2000.
\newblock Headline generation based on statistical translation.
\newblock \emph{ACL}, 318--325.

\bibitem[{Bowman et~al.(2015)Bowman, Angeli, Potts, and
  Manning}]{bowman2015large}
Bowman, S.; Angeli, G.; Potts, C.; and Manning, C.~D. 2015.
\newblock A large annotated corpus for learning natural language inference.
\newblock In \emph{EMNLP}, 632--642.

\bibitem[{Brown et~al.(2020)Brown, Mann, Ryder, Subbiah, Kaplan, Dhariwal,
  Neelakantan, Shyam, Sastry, Askell et~al.}]{brown2020language}
Brown, T.~B.; Mann, B.; Ryder, N.; Subbiah, M.; Kaplan, J.; Dhariwal, P.;
  Neelakantan, A.; Shyam, P.; Sastry, G.; Askell, A.; et~al. 2020.
\newblock Language models are few-shot learners.
\newblock \emph{arXiv preprint arXiv:2005.14165}.

\bibitem[{Chen et~al.(2019)Chen, Lin, Zhang, Yang, Zhou, and
  Tang}]{chen2019towards}
Chen, Q.; Lin, J.; Zhang, Y.; Yang, H.; Zhou, J.; and Tang, J. 2019.
\newblock Towards knowledge-based personalized product description generation
  in e-commerce.
\newblock In \emph{KDD}, 3040--3050.

\bibitem[{Chen et~al.(2021)Chen, Li, Yu, Wu, and Miao}]{Chen-et-al:2021HyDRA}
Chen, Y.; Li, B.; Yu, H.; Wu, P.; and Miao, C. 2021.
\newblock HyDRA: Hypergradient Data Relevance Analysis for Interpreting Deep
  Neural Networks.
\newblock In \emph{AAAI}, 7081--7089.

\bibitem[{Devlin et~al.(2019{\natexlab{a}})Devlin, Chang, Lee, and
  Toutanova}]{devlin-etal-2019-bert}
Devlin, J.; Chang, M.-W.; Lee, K.; and Toutanova, K. 2019{\natexlab{a}}.
\newblock {BERT}: Pre-training of Deep Bidirectional Transformers for Language
  Understanding.
\newblock In \emph{NAACL}, 4171--4186.

\bibitem[{Devlin et~al.(2019{\natexlab{b}})Devlin, Chang, Lee, and
  Toutanova}]{devlin2019bert}
Devlin, J.; Chang, M.-W.; Lee, K.; and Toutanova, K. 2019{\natexlab{b}}.
\newblock BERT: Pre-training of Deep Bidirectional Transformers for Language
  Understanding.
\newblock In \emph{ACL}.

\bibitem[{Dong et~al.(2019)Dong, Yang, Wang, Wei, Liu, Wang, Gao, Zhou, and
  Hon}]{dong2019unified}
Dong, L.; Yang, N.; Wang, W.; Wei, F.; Liu, X.; Wang, Y.; Gao, J.; Zhou, M.;
  and Hon, H.-W. 2019.
\newblock Unified Language Model Pre-training for Natural Language
  Understanding and Generation.
\newblock arXiv:1905.03197.

\bibitem[{Gehring et~al.(2017)Gehring, Auli, Grangier, Yarats, and
  Dauphin}]{gehring2017convolutional}
Gehring, J.; Auli, M.; Grangier, D.; Yarats, D.; and Dauphin, Y.~N. 2017.
\newblock Convolutional Sequence to Sequence Learning.
\newblock arXiv:1705.03122.

\bibitem[{Keskar et~al.(2019)Keskar, McCann, Varshney, Xiong, and
  Socher}]{keskar2019ctrl}
Keskar, N.~S.; McCann, B.; Varshney, L.~R.; Xiong, C.; and Socher, R. 2019.
\newblock CTRL: A Conditional Transformer Language Model for Controllable
  Generation.
\newblock arXiv:1909.05858.

\bibitem[{Knight and Marcu(2000)}]{kevin2000statistics}
Knight, K.; and Marcu, D. 2000.
\newblock Statistics-based summarization-step one: Sentence compression.
\newblock \emph{AAAI}, 703--710.

\bibitem[{Lafferty, McCallum, and Pereira(2001)}]{laffertyCrf}
Lafferty, J.~D.; McCallum, A.; and Pereira, F. C.~N. 2001.
\newblock Conditional Random Fields: Probabilistic Models for Segmenting and
  Labeling Sequence Data.
\newblock In \emph{ICML}, 282--289. San Francisco, CA, USA.

\bibitem[{Lavie, Sagae, and Jayaraman(2004)}]{lavie2004significance}
Lavie, A.; Sagae, K.; and Jayaraman, S. 2004.
\newblock The significance of recall in automatic metrics for MT evaluation.
\newblock In \emph{AMTA}, 134--143.

\bibitem[{Lewis et~al.(2019)Lewis, Liu, Goyal, Ghazvininejad, Mohamed, Levy,
  Stoyanov, and Zettlemoyer}]{lewis2019bart}
Lewis, M.; Liu, Y.; Goyal, N.; Ghazvininejad, M.; Mohamed, A.; Levy, O.;
  Stoyanov, V.; and Zettlemoyer, L. 2019.
\newblock Bart: Denoising sequence-to-sequence pre-training for natural
  language generation, translation, and comprehension.
\newblock \emph{arXiv preprint arXiv:1910.13461}.

\bibitem[{Li et~al.(2020)Li, Yuan, Xu, Wu, He, and Zhou}]{li2020aspect}
Li, H.; Yuan, P.; Xu, S.; Wu, Y.; He, X.; and Zhou, B. 2020.
\newblock Aspect-aware multimodal summarization for chinese e-commerce
  products.
\newblock In \emph{AAAI}, volume~34, 8188--8195.

\bibitem[{Li et~al.(2018)Li, Song, Zhang, Chen, Shi, Zhao, and
  Yan}]{li-etal-2018-generating-classical}
Li, J.; Song, Y.; Zhang, H.; Chen, D.; Shi, S.; Zhao, D.; and Yan, R. 2018.
\newblock Generating Classical {C}hinese Poems via Conditional Variational
  Autoencoder and Adversarial Training.
\newblock In \emph{EMNLP}, 3890--3900.

\bibitem[{Lin(2004)}]{lin2004rouge}
Lin, C.-Y. 2004.
\newblock Rouge: A package for automatic evaluation of summaries.
\newblock In \emph{Text summarization branches out}, 74--81.

\bibitem[{Liu et~al.(2019)Liu, Ott, Goyal, Du, Joshi, Chen, Levy, Lewis,
  Zettlemoyer, and Stoyanov}]{liu2019roberta}
Liu, Y.; Ott, M.; Goyal, N.; Du, J.; Joshi, M.; Chen, D.; Levy, O.; Lewis, M.;
  Zettlemoyer, L.; and Stoyanov, V. 2019.
\newblock Roberta: A robustly optimized bert pretraining approach.
\newblock In \emph{ACL}.

\bibitem[{Och et~al.(2004)Och, Gildea, Khudanpur, Sarkar, Yamada, Fraser,
  Kumar, Shen, Smith, Eng, Jain, Jin, and Radev}]{och-etal-2004-smorgasbord}
Och, F.~J.; Gildea, D.; Khudanpur, S.; Sarkar, A.; Yamada, K.; Fraser, A.;
  Kumar, S.; Shen, L.; Smith, D.; Eng, K.; Jain, V.; Jin, Z.; and Radev, D.
  2004.
\newblock A Smorgasbord of Features for Statistical Machine Translation.
\newblock In \emph{NAACL-HLT}, 161--168.

\bibitem[{Papineni et~al.(2002)Papineni, Roukos, Ward, and
  Zhu}]{papineni2002bleu}
Papineni, K.; Roukos, S.; Ward, T.; and Zhu, W.-J. 2002.
\newblock Bleu: a method for automatic evaluation of machine translation.
\newblock In \emph{ACL}, 311--318.

\bibitem[{Post(2018)}]{post2018call}
Post, M. 2018.
\newblock A call for clarity in reporting BLEU scores.
\newblock \emph{arXiv preprint arXiv:1804.08771}.

\bibitem[{Radford et~al.(2018{\natexlab{a}})Radford, Narasimhan, Salimans, and
  Sutskever}]{radfordimproving}
Radford, A.; Narasimhan, K.; Salimans, T.; and Sutskever, I.
  2018{\natexlab{a}}.
\newblock Improving language understanding by generative pre-training.
\newblock \emph{URL https://s3-us-west-2. amazonaws.
  com/openai-assets/researchcovers/languageunsupervised/language understanding
  paper. pdf}.

\bibitem[{Radford et~al.(2018{\natexlab{b}})Radford, Wu, Child, Luan, Amodei,
  and Sutskever}]{radford2019language}
Radford, A.; Wu, J.; Child, R.; Luan, D.; Amodei, D.; and Sutskever, I.
  2018{\natexlab{b}}.
\newblock Language Models are Unsupervised Multitask Learners.

\bibitem[{Radford et~al.(2019)Radford, Wu, Child, Luan, Amodei, and
  Sutskever}]{radford:2019:arxiv}
Radford, A.; Wu, J.; Child, R.; Luan, D.; Amodei, D.; and Sutskever, I. 2019.
\newblock Language models are unsupervised multitask learners.
\newblock \emph{OpenAI Blog}, 1(8).

\bibitem[{Rajpurkar et~al.(2016)Rajpurkar, Zhang, Lopyrev, and
  Liang}]{rajpurkar2016squad}
Rajpurkar, P.; Zhang, J.; Lopyrev, K.; and Liang, P. 2016.
\newblock SQuAD: 100,000+ Questions for Machine Comprehension of Text.
\newblock In \emph{EMNLP}, 2383--2392.

\bibitem[{Reiter and Dale(2000)}]{reiter_dale_2000}
Reiter, E.; and Dale, R. 2000.
\newblock \emph{Building Natural Language Generation Systems}.
\newblock Studies in Natural Language Processing. Cambridge University Press.

\bibitem[{Rush, Chopra, and Weston(2015)}]{rush2015neural}
Rush, A.~M.; Chopra, S.; and Weston, J. 2015.
\newblock A neural attention model for abstractive sentence summarization.
\newblock \emph{EMNLP}, 379--389.

\bibitem[{Song et~al.(2019)Song, Tan, Qin, Lu, and Liu}]{song2019mass}
Song, K.; Tan, X.; Qin, T.; Lu, J.; and Liu, T.-Y. 2019.
\newblock MASS: Masked Sequence to Sequence Pre-training for Language
  Generation.
\newblock In \emph{ICML}, 5926--5936. PMLR.

\bibitem[{Su(2021)}]{zhuiyit5pegasus}
Su, J. 2021.
\newblock T5 PEGASUS - ZhuiyiAI.
\newblock Technical report.

\bibitem[{Sutskever, Vinyals, and Le(2014)}]{sutskever2014sequence}
Sutskever, I.; Vinyals, O.; and Le, Q.~V. 2014.
\newblock Sequence to Sequence Learning with Neural Networks.
\newblock arXiv:1409.3215.

\bibitem[{Tao et~al.(2018)Tao, Gao, Shang, Wu, Zhao, and Yan}]{ijcai2018-614}
Tao, C.; Gao, S.; Shang, M.; Wu, W.; Zhao, D.; and Yan, R. 2018.
\newblock Get The Point of My Utterance! Learning Towards Effective Responses
  with Multi-Head Attention Mechanism.
\newblock In \emph{IJCAI}, 4418--4424.

\bibitem[{Vaswani et~al.(2017)Vaswani, Shazeer, Parmar, Uszkoreit, Jones,
  Gomez, Kaiser, and Polosukhin}]{vaswani2017attention}
Vaswani, A.; Shazeer, N.; Parmar, N.; Uszkoreit, J.; Jones, L.; Gomez, A.~N.;
  Kaiser, L.; and Polosukhin, I. 2017.
\newblock Attention Is All You Need.
\newblock arXiv:1706.03762.

\bibitem[{Wang et~al.(2017)Wang, Hou, Liu, Cao, and Lin}]{wang2017statistical}
Wang, J.; Hou, Y.; Liu, J.; Cao, Y.; and Lin, C.-Y. 2017.
\newblock A statistical framework for product description generation.
\newblock In \emph{IJCNLP}, 187--192.

\bibitem[{Wu et~al.(2020)Wu, Pan, Chen, Long, Zhang, and
  Philip}]{wu2020comprehensive}
Wu, Z.; Pan, S.; Chen, F.; Long, G.; Zhang, C.; and Philip, S.~Y. 2020.
\newblock A comprehensive survey on graph neural networks.
\newblock \emph{IEEE Transactions on Neural Networks and Learning Systems},
  32(1): 4--24.

\bibitem[{Yao et~al.(2019)Yao, Peng, Weischedel, Knight, Zhao, and
  Yan}]{Yao_Peng_Weischedel_Knight_Zhao_Yan_2019}
Yao, L.; Peng, N.; Weischedel, R.; Knight, K.; Zhao, D.; and Yan, R. 2019.
\newblock Plan-and-Write: Towards Better Automatic Storytelling.
\newblock \emph{AAAI}, 33(01): 7378--7385.

\bibitem[{Zhan et~al.(2021)Zhan, Zhang, Chen, Shen, Ding, Bao, Yan, and
  Lan}]{zhan2021probing}
Zhan, H.; Zhang, H.; Chen, H.; Shen, L.; Ding, Z.; Bao, Y.; Yan, W.; and Lan,
  Y. 2021.
\newblock Probing Product Description Generation via Posterior Distillation.
\newblock In \emph{AAAI}, volume~35, 14301--14309.

\bibitem[{Zhang et~al.(2021)Zhang, Jiang, Shang, Cheng, Zhang, Fan, Xiao, and
  Long}]{zhang2021dsgpt}
Zhang, X.; Jiang, Y.; Shang, Y.; Cheng, Z.; Zhang, C.; Fan, X.; Xiao, Y.; and
  Long, B. 2021.
\newblock DSGPT: Domain-Specific Generative Pre-Training of Transformers for
  Text Generation in E-commerce Title and Review Summarization.
\newblock In \emph{SIGIR}, 2146–2150.

\bibitem[{Zhang, Zhao, and LeCun(2015)}]{zhang2015character}
Zhang, X.; Zhao, J.; and LeCun, Y. 2015.
\newblock Character-level convolutional networks for text classification.
\newblock \emph{NeurIPS}, 28: 649--657.

\bibitem[{Zhang et~al.(2020)Zhang, Sun, Galley, Chen, Brockett, Gao, Gao, Liu,
  and Dolan}]{zhang2020dialogpt}
Zhang, Y.; Sun, S.; Galley, M.; Chen, Y.-C.; Brockett, C.; Gao, X.; Gao, J.;
  Liu, J.; and Dolan, W.~B. 2020.
\newblock DIALOGPT: Large-Scale Generative Pre-training for Conversational
  Response Generation.
\newblock In \emph{ACL}, 270--278.

\bibitem[{Zou et~al.(2020)Zou, Zhang, Lu, Wei, and Zhou}]{zou2020pre}
Zou, Y.; Zhang, X.; Lu, W.; Wei, F.; and Zhou, M. 2020.
\newblock Pre-training for Abstractive Document Summarization by Reinstating
  Source Text.
\newblock In \emph{EMNLP}, 3646--3660.

\end{thebibliography}

\end{document}